\documentclass[letterpaper]{article}
\usepackage{aaai}
\usepackage{times}
\usepackage{graphicx}
\usepackage{latexsym}
\usepackage{epic}
\usepackage{amsfonts}
\usepackage{amsmath}
\usepackage{amssymb}
\usepackage{amsthm}
\usepackage{xspace}
\usepackage{mathrsfs}
\usepackage{subfigure}

\newcommand{\mf}{{\mathscr F}}
\newcommand{\mc}{\mathcal C}
\newcommand{\ra}{\rightarrow}

\newcommand{\ms}{\mathcal S}
\newcommand{\ma}{\mathcal A}
\newcommand{\mv}{\mathcal V}
\newcommand{\rev}{\text{rev}}
\newcommand{\others}{\text{\it Others}}

\begin{document}

\newtheorem{theorem}{Theorem}
\newtheorem{lemma}{Lemma}
\newtheorem{definition}{Definition}
\newtheorem{myexample}{Example}
\newtheorem{mytheorem}{Theorem}
\newcommand{\myproof}{\noindent {\bf Proof:\ \ }}
\newcommand{\sketch}{\noindent {\bf Proof sketch:\ \ }}
\newcommand{\myqed}{\mbox{$\Box$}}
\newcommand{\nmax}{N}
\newcommand{\zuck}{\mbox{\sc Reverse}}
\newcommand{\lslg}{\mbox{\sc Largest Fit}}
\newcommand{\lsla}{\mbox{\sc Average Fit}}
\newcommand{\lslgbf}{\mbox{\bf Largest Fit}}
\newcommand{\lslabf}{\mbox{\bf Average Fit}}
\newcommand{\myOmit}[1]{}
\newcommand{\vote}[3]{\mbox{$#1 \! \succ \! #2 \! \succ \! #3$}\xspace}
\newcommand{\vvote}[4]{\mbox{$#1 \! \succ \! #2 \! \succ \! #3 \! \succ \! #4$}\xspace}

\title{Manipulation of Nanson's and Baldwin's Rules}

\author{Nina Narodytska\\
NICTA and UNSW\\ Sydney, Australia\\
ninan@cse.unsw.edu.au \And Toby Walsh\\
NICTA and UNSW\\ Sydney, Australia\\
toby.walsh@nicta.com.au
 \And Lirong Xia\\ Department of Computer Science\\ Duke University \\ Durham, NC 27708, USA\\
lxia@cs.duke.edu}

\maketitle

\begin{abstract}
Nanson's and Baldwin's voting rules select
a winner by successively eliminating
candidates with low Borda scores. We show
that these rules have a number of desirable
computational properties. In particular,
with unweighted votes, it is
NP-hard to manipulate either rule with one manipulator, whilst
with weighted votes, it is
NP-hard to manipulate either rule with a small number of
candidates and a coalition of manipulators.
As only a couple of other voting rules
are known to be NP-hard to manipulate
with a single manipulator, Nanson's
and Baldwin's rules appear
to be particularly resistant to manipulation from a theoretical perspective.
We also propose a number of approximation methods
for manipulating these two rules.
Experiments demonstrate that both rules are
often difficult to manipulate in practice.
These results suggest that elimination style
voting rules deserve further study.
\end{abstract}

\section{Introduction}

Computational social choice studies computational aspects of voting.
For example, how does a coalition of agents compute a manipulation?
Can we compile these votes into a more compact form? How do we
decide if we have elicited enough votes from the agents to be able
to declare the result? Whilst there has been a very active research
community studying these sort of questions for well known voting
rules like plurality and Borda, there are other less well known
rules that might deserve attention. In particular, we put
forward two historical voting rules due to Nanson and Baldwin which
are related to Borda voting.

There are several reasons to consider these two rules. Firstly, they
have features that might appeal to the two opposing camps that
support Borda and Condorcet. In particular, both rules are Condorcet
consistent as they elect the candidate who beats all others in
pairwise elections.
Secondly, both rules are elimination style procedures where
candidates are successively removed. Other elimination procedures
like STV and plurality with runoff are computationally hard to
manipulate (in the case of STV, with or without weights on the
votes, whilst in the case of plurality with runoff, only in the case
of weighted votes). We might therefore expect Nanson's and Baldwin's
rules to be computationally hard to manipulate. Thirdly, statistical
analysis suggest that, whilst the Borda rule is vulnerable to
manipulation \cite{chamberlin}, Nanson's rule is particularly
resistant \cite{favardin}. We might expect Baldwin to be similarly
resistant. Finally, the two rules have been used in real elections in
the Universitiy of Melbourne (between 1926 and 1982),
the University of Adelaide (since 1968), and the State
of Michigan (in the 1920s).
It is perhaps therefore somewhat surprising that
neither rule has received much attention till now in the
computational social choice literature.

\section{Preliminaries}
Let $\mc=\{c_1,\ldots,c_m\}$ be the set of {\em candidates} (or {\em
alternatives}).  A linear order on $\mc$ is a transitive,
antisymmetric, and total relation on $\mc$.  The set of all linear
orders on $\mc$ is denoted by $L(\mc)$.  An $n$-voter profile $P$ on
$\mc$ consists of $n$ linear orders on $\mc$.  That is,
$P=(V_1,\ldots,V_n)$, where for every $j\leq n$, $V_j\in L(\mc)$.
The set of all $n$-profiles is denoted by $\mf_n$. We let $m$ denote
the number of candidates. A (deterministic) {\em voting rule} $r$ is
a function that maps any profile on $\mc$ to a unique winning
candidate, that is, $r:\mf_1\cup\mf_2\cup\ldots\ra \mc$. In this
paper, if not mentioned otherwise, ties are broken in the fixed
order $c_1\succ c_2\succ\cdots\succ c_m$.

{\em (Positional) scoring rules} are commonly
used voting rules. Each positional scoring rule is identified by a
{\em scoring vector}
  $\vec s_m=(\vec s_m(1),\ldots,\vec s_m(m))$ of $m$ integers, for any vote $V\in L(\mc)$ and any candidate $c\in
  \mc$, let $\vec s_m(c,V)=\vec s_m(j)$, where $j$ is the rank of $c$ in $V$.  For any
  profile $P=(V_1,\ldots,V_n)$, let $\vec s_m(c,P)=\sum_{j=1}^n\vec s_m(c,V_j)$.
  The rule selects $c\in \mc$ such that the total score $\vec s_m(c,P)$ is
  maximized.
We assume scores are integers and decreasing. {\em Borda} is the
positional scoring rule that corresponds to the scoring vector
$(m-1, m-2, \ldots,0)$. We write $s(a,P)$ for the Borda score given
to candidate $a$ from the profile of votes $P$, and $s(a)$ where $P$
is obvious from the context.
When voters are weighted (that is, each voter is associated with a
positive real number as the weight), a positional scoring rule
selects the candidate that maximizes the weighted total score.

The \emph{unweighted (coalitional) manipulation} problem is defined
as follows. An instance is a tuple $(r,P^{NM},c,M)$, where $r$ is a
voting rule, $P^{NM}$ is the non-manipulators' profile, $c$ is the
candidate preferred by the manipulators, and $M$ is the set of
manipulators. We are asked whether there exists a profile $P^{M}$
for the manipulators such that $r(P^{NM}\cup P^{M})=c$. The {\em
weighted (coalitional) manipulation} is defined similarly, where the
weights of the voters (both non-manipulators and manipulators) are
also given as inputs. As is common in the literature, we break
ties in favour of the coalition of the manipulators where appropriate.

\section{Nanson's and Baldwin's Rules}

The Borda rule has several good properties. For instance, it is
monotonic as increasing the score for a candidate only helps them
win. Also it never elects the Condorcet loser (a candidate that
loses to all others in a majority of head to head elections).
However, it may not elect the Condorcet winner (a candidate that
beats all others in a majority of head to head elections).
Nanson's and Baldwin's rules, by comparison, always elect
the Condorcet winner when it exists.

\myOmit{
We will number candidates from 1 to $m$.
We suppose a coalition of $n$ agents
are collectively trying to ensure a preferred candidate $d$ wins.
We let $s(i)$ be the score candidate $i$ receives from
the votes cast so far. A \emph{score vector} $\langle s(1), \ldots,
s(n)\rangle$ gives the scores of the candidates from a
set of votes.
Given a set of votes, we define the \emph{gap} of candidate $i$
as $g(i) = s(d) + n(m-1) - s(i)$.
For $d$
to win, we need additional votes that
add a score to candidate $i$ which is less than
or equal to $g(i)$. Note that if $g(i)$ is negative for any $i$,
then $d$ cannot win and manipulation is impossible.
}


Nanson's and Baldwin's rules are derived from the Borda rule.
{Nanson's rule} 
eliminates all those candidates with less than the average Borda
score~\cite{nanson}. The rule is then repeated with the reduced set
of candidates until there is a single candidate left. A closely
related voting
rule proposed by 
Baldwin successively eliminates the candidate with the lowest Borda
score\footnote{If multiple candidates have the lowest score, then we
use a tie-breaking mechanism to eliminate one of them.} until one
candidate remains \cite{baldwin}.
The two rules are closely related, and
indeed are sometimes confused. 
%
%
One of the most appealing properties of
Nanson's and Baldwin's rules is that they
are Condorcet consistent, i.e. they elect the Condorcet
winner. 
This follows from the fact that the Borda score of the Condorcet
winner is never below the average Borda score. Both rules possess
several other desirable properties including the majority criterion
and the Condorcet loser criterion. There are also properties which
distinguish them apart. For instance, Nanson's rule satisfies
reversal symmetry (i.e.~if there is a unique winner and voters
reverse their vote then the winner changes) but Baldwin's rule does
not.

\section{Unweighted Manipulation}

We start by considering the
computational complexity of manipulating both
these rules with unweighted votes. 
We prove that the coalitional manipulation problem is NP-complete
for both rules even with a single manipulator. Computational
intractability with a single manipulator is known only for a small
number of other voting rules including
the second order Copeland rule~\cite{btt},
STV~\cite{Bartholdi91:Single}
and ranked pairs~\cite{Xia09:Complexity}. In contrast, when there
are two or more manipulators, unweighted coalitional manipulation is
hard for some other common voting
rules~\cite{Faliszewski08:Copeland,Faliszewski10:Manipulation,Xia10:Scheduling,dknwaaai11,Betzler11:Unweighted}.
Our results therefore significantly
increase the size of the set of voting rules used in practice that
are known to be NP-hard to manipulate with a single manipulator.
This also contrasts to Borda where computing a manipulation with a
single manipulator is polynomial~\cite{btt}. Adding elimination
rounds to Borda to get Nanson's or Baldwin's rules increases the
computational complexity of computing a manipulation with one
manipulator from polynomial to NP-hard.

Our results are proved by reductions from  the {\sc exact 3-cover
(x3c)} problem. An {\sc x3c} instance contains two sets:
$\mv=\{v_1,\ldots,v_q\}$ and $\ms=\{S_1,\ldots,S_t\}$, where $t\geq
2$ and for all $j\leq t$, $|S_j|=3$ and $S_j\subseteq \mv$. We are
asked whether there exists a subset $\ms'$ of $\ms$ such that each
element in $\mv$ is in exactly one of the 3-sets in $\ms'$.
\begin{mytheorem}
\label{thm:Baldwin}
With unweighted votes, the coalitional manipulation problem under Baldwin's rule
is NP-complete even when there is only one manipulator.
\end{mytheorem}
\myproof We sketch a reduction from  {\sc x3c}. Given an {\sc x3c}
instance $\mv=\{v_1,\ldots,v_q\}, \ms=\{S_1,\ldots,S_t\}$, we let
the set of candidates be $\mc=\{c,d,b\}\cup \mv\cup \ma$,
where $c$ is the candidate that the manipulator wants to make the winner, 
$\ma = \{a_1,\ldots,a_t\}$, and $d$ and $b$ are additional
candidates. Members of $\ma$ correspond to the 3-sets in $\ms$. Let
$m=|\mc|=q+t+3$.

The profile $P$ contains two parts: $P_1$, which is used to
control the changes in the score differences between candidates,
after a set of candidates are removed, and $P_2$, which is used to
balance the score differences between the candidates.
We define the votes $W_{(u,v)}=\{\vote{u}{v}{{\others}},
\vote{\rev({\others})}{u}{v}\}$ where $\others$ is a total order in
which the candidates in $\mc\setminus\{u,v\}$ are in a pre-defined
lexicographic order, and $\rev({\others})$ is the reverse.

We make the following observations
on $W_{(c_1,c_2)}$. For any set of candidates $\mc'\subseteq \mc$
and any pair of candidates $e_1,e_2\in \mc\setminus \mc'$,
\begin{align*}
&s(e_1,W_{(c_1,c_2)}|_{\mc\setminus\mc'})-s(e_2,W_{(c_1,c_2)}|_{\mc\setminus\mc'})\\
=\ &
s(e_1,W_{(c_1,c_2)})-s(e_2,W_{(c_1,c_2)})\\&\hspace{1cm}+\left\{\begin{array}
{ll}1&\text{if } e_1=c_2\text{ and }c_1\in \mc'\\
-1&\text{if }e_1=c_1\text{ and }c_2\in \mc'\\
0&\text{otherwise}
\end{array}\right.
\end{align*}
Here $W_{(c_1,c_2)}|_{\mc\setminus\mc'}$ is the pair of votes
obtained from $W$ by removing all candidates in $\mc'$. In words,
the formula states that after $\mc'$ is removed, the score
difference between $e_1$ and $e_2$ is increased by $1$ if and only
if $e_1=c_2$ and $c_1$ is removed; it is decreased by $1$ if and
only if $e_1=c_1$ and $c_2$ is removed; for any other cases, the
score difference does not change. Moreover, for any $e\in
\mc\setminus \{c_1,c_2\}$,
$s(c_1,W_{(c_1,c_2)})-s(e,W_{(c_1,c_2)})=1$ and
$s(c_2,W_{(c_1,c_2)})-s(e,W_{(c_1,c_2)})=-1$.

We next show how to use $W_{(c_1,c_2)}$ to construct the first part
of the profile $P_1$. Let $m=|\mc|$, that is, $m=q+t+3$. $P_1$ is
composed of the following votes: (1) for each $j\leq t$ and each
$v_i\in S_j$, there are $2m$ copies of $W_{(v_i,a_j)}$; (2) for each
$i\leq q$, there are $m$ copies of $W_{(b,v_i)}$; (3) there are
$m(t+6)$ copies of $W_{(b,c)}$. It is not hard to verify that
$s(b,P_1)-s(c,P_1)\geq mq$, and for any $c'\in \mv\cup \ma$,
$s(c',P_1)-s(c,P_1)\geq 2m$. $P_2$ is composed of the following
votes: (1) for each $i\leq q$, there are $s(v_i,P_1)-s(c,P_1)-m$
copies of $W_{(d,v_i)}$; (2) for each $j\leq t$, there are
$s(a_j,P_1)-s(c,P_1)-1$ copies of $W_{(d,a_j)}$; (3) there are
$s(b,P_1)-s(c,P_1)-mq$ copies of $W_{(d,b)}$.

Let $P=P_1\cup P_2$. We make the following observations on the Borda scores of the candidates in $P$.

$\bullet$ For any $i\leq q$, $s(v_i,P)-s(c,P)=m$;

$\bullet$  for any $j\leq t$, $s(a_j,P)-s(c,P)=1$;

$\bullet$  $s(b,P)-s(c,P)=mq$.

Suppose the {\sc x3c} instance has a solution, denoted by (after
reordering the sets in $\ms$)
$S_{1},\ldots, S_{q/3}$. Then, we let the manipulator vote for:

$c\succ d\succ a_{q/3+1}\succ\cdots\succ a_{t}\succ b\succ \mv\succ
a_1\succ\cdots\succ a_{q/3}$

In the first $4q/3$ rounds, all candidates in $\mv$ and
$\{a_1,\ldots,a_{q/3}\}$ drop out. Then $b$ drops out. In the
following $t-q/3$ rounds the candidates in
$\{a_{q/3+1},\ldots,a_{t}\}$ drop out. Finally, $d$ loses to $c$ in
their pairwise election, which means that $c$ is the winner.

Suppose the manipulator can cast a vote to make $c$ the winner. We
first note that $d$ must be eliminated in the final round since its
score is higher than $c$ in all previous rounds.  In the round when
$b$ is eliminated, the score of $b$ should be no more than the score
of $c$. We note that $s(b,P)-s(c,P)=mq$ and the score difference can
only be reduced by the manipulator ranking $b$ below $c$, and by
eliminating $v_1,\ldots,v_q$ before $b$. However, by ranking $b$
below $c$, the score difference is reduced by no more than $m-1$.
Therefore, before $b$ drops out, all candidates in $\mv$ must have
already dropped out. We note that for any $v_i\in \mv$,
$s(v_i,P)-s(c,P)=m$. Therefore, for each $v_i\in \mv$, there exists
$a_j$ with $v_i\in S_j$ who is removed before $v_i$. For any such
$a_j$, none of the candidates in $S_j$ can drop out before $a_j$
(otherwise the score of $a_j$ cannot be less than $c$ before $b$
drops out), and in the next three rounds the candidates in $S_j$
drop out. It follows that the set of candidates in $\ma$ that drop
out before any candidate in $\mv$ corresponds to an exact cover of
$\mv$.
%
\hfill\myqed

\begin{mytheorem}
With unweighted votes, the coalitional manipulation problem under Nanson's rule
is NP-complete even when there is only one manipulator.
\end{mytheorem}

The proof uses the same gadget $W_{(u,v)}$ that is used in
the proof of Theorem~\ref{thm:Baldwin}. Due to the space
constraints, the proof can be found in an online technical
report.

\myOmit{\myproof The idea of the proof is similar to the proof of
Theorem~\ref{thm:Baldwin}. We again use a reduction from {\sc x3c}.
Given an {\sc x3c} instance $\mv=\{v_1,\ldots,v_q\}$, $\ms =
\{S_1,\ldots,S_t\}$, we let the set of candidates be
$\mc=\{c,d,b_1,b_2\}\cup \mv\cup\ma$, where
$d$, $b_1$,
and $b_2$ are additional candidates. Without loss of generality,
we suppose $q$ and $t$ are even, and $t\geq 3q$.  We will use the votes
$W_{(u,v)}$ defined in the last proof 
to
construct the profile.

Let $m=|\mc|=q+t+4$. Again, the profile has two parts: $P_1$, which
is used to control the score differences between the candidates and
the average score, and $P_2$, which is used to set the initial
scores. $P_1$ is composed of the following votes: (1) for each
$j\leq t$, there are $7m/2-q/3$ copies of $W_{(a_j,b_1)}$, and for
each $v_i\in S_j$, there are $m$ copies of $W_{(v_i,a_j)}$; (2) for
each $i\leq q$, there are $m$ copies of $W_{(v_i,c)}$; (3) there are
$mq$ copies of $W_{(c,b_1)}$ and $mq+t(7m/2-q/3)$ copies of
$W_{(b_1,b_2)}$. The second part of the profile $P_2$ is composed of
the following votes: for each $i\leq q$, there are
$m\cdot\Delta(v_i)$ copies of $W_{(d,v_i)}$, where $\Delta(v_i)$ is
the number of times that $v_i$ is covered by the 3-sets in $\ms$.

Suppose the {\sc x3c} instance has a solution, denoted by
$S_{1},\ldots, S_{q/3}$. Then, we let the
manipulator vote: $$c\succ b_1\succ b_2\succ
d\succ a_{q/3+1}\succ\cdots\succ a_{t}\succ \mv\succ
a_1\succ\cdots\succ a_{q/3}$$ In the first round
$b_2, a_1,\ldots,a_{q/3}$ are eliminated; in the second round, the candidates
in $\mv$ are eliminated; in the following rounds, $a_{q/3+1},\ldots,a_t,d$
are eliminated and $c$ wins.

Suppose the manipulator can cast a vote to make $c$ the winner. It
is easy to check that $b_2$ definitely drops out in the first round,
$b_1$ in the second round, and $d$ remains till the final round. In
the third round if any $v_i$ still remains, then the score of $c$
will be strictly lower than the average score. Therefore, all
candidates in $\mv$ must drop out in the first two rounds. We note
that $v_i$ can only drop out in the second round if a candidate
$a_j$ with $v_i\in S_j$ drops out in the first round. Moreover, no
more than $q/3$ candidates in $\ma$ can possibly drop out in the
first round (the only way to eliminate $a\in \ma$ in the first round
is by ranking it in the bottom $q/3$ positions). Therefore, in order
for $c$ to survive beyond the third round, the bottom $q/3$
candidates in the manipulator's vote must be among $\ma$ and they
must correspond to an exact cover of $\mv$.
%
\hfill\myqed }

\section*{Weighted Manipulation}

If the number of candidates is bounded, then manipulation is NP-hard
to compute when votes are weighted. Baldwin's rule appears more
computationally difficult than Nanson's rule.
\citeauthor{Coleman07:Complexity}
\shortcite{Coleman07:Complexity}
prove that Baldwin's requires only 3 candidates to be
NP-hard, whilst we prove here that
Nanson's rule is polynomial to manipulate with
3 candidates and requires at least 4 candidates to
be NP-hard. It follows that computing a
manipulation is NP-hard for both rules when votes are unweighted,
the number of candidates is small and there is uncertainty about how
agents have voted in the form of a probability distribution
\cite{csljacm2007}. Note that the coalition manipulation problem for
Borda with weighted votes
is NP-hard for 3 or more candidates \cite{csljacm2007}. Thus,
somewhat surprisingly, adding an elimination round to Borda, which
gives us Nanson's rule, decreases the computational complexity of
computing a manipulation with 3 manipulators from NP-hard to
polynomial.

\begin{mytheorem}
\label{t:baldwin_constructive}
With Nanson's rule and weighted votes,
the coalition manipulation problem is NP-complete for just 4 candidates.
\end{mytheorem}
\myproof The proof is by a reduction from {\sc partition}, where we
are given a group of integers $\{k_1,\ldots,k_l\}$ with sum $2K$,
and we are asked whether there is way to partition the group into
two groups, the elements in each of which sum to $K$. For any {\sc
partition} instance, we construct a coalition manipulation problem
with 4 candidates ($a$, $b$, $c$ and $p$) where $p$ is again the
candidate that the manipulators wish to win. \myOmit{
Table~\ref{t:nonman} shows the votes of non-manipulators. The
preferred candidate is $a$.
\begin{table}
\label{t:nonman}
\center{
\begin{tabular}{|c|c|c|c|c|c|c|c|}
  \hline
  weights & \multicolumn{2}{|c|}{$1^{st}$}&  \multicolumn{2}{|c|}{$2^{nd}$ }& \multicolumn{2}{|c|}{$3^{rd}$} &  {$4^{th}$}  \\
    \hline
    \multicolumn {8}{|c|}{Non-manipulators votes}\\
    \hline
  $2K + 1$& $b $&$\succ$& $a$ &$\succ$& $c$ & $\succ$ & $ d$ \\
  $2K + 1$& $d $&$\succ$& $c$ &$\succ$& $b$ & $\succ$ & $ a$ \\
  $2K + 1$& $c $&$\succ$& $a$ &$\succ$& $b$ & $\succ$ & $ d$ \\
  $2K + 1$& $d $&$\succ$& $b$ &$\succ$& $c$ & $\succ$ & $ a$ \\
  $K + 2$& $a $&$\succ$& $d$ &$\succ$& $b$ & $\succ$ & $ c$ \\
  $K + 2$& $c $&$\succ$& $b$ &$\succ$& $a$ & $\succ$ & $ d$ \\
  $1$& $d $&$\succ$& $b$ &$\succ$& $a$ & $\succ$ & $ c$ \\
  $1$& $c $&$\succ$& $a$ &$\succ$& $d$ & $\succ$ & $ b$ \\
  $1$& $d $&$\succ$& $c$ &$\succ$& $a$ & $\succ$ & $ b$ \\
  $1$& $b $&$\succ$& $a$ &$\succ$& $d$ & $\succ$ & $ c$ \\

  \hline
\end{tabular}
\caption{The scores of non-manipulators}
}
\end{table}
}
We suppose the non-manipulators
have voted as follows:
$2K + 1$ for each of
\vvote{b}{p}{c}{a}, 
\vvote{a}{c}{b}{p}, 
\vvote{c}{p}{b}{a} and 
\vvote{a}{b}{c}{p}, 
$K + 2$ for
\vvote{p}{a}{b}{c} and 
\vvote{c}{b}{p}{a}, and 
1 each for
\vvote{a}{b}{p}{c}, 
\vvote{c}{p}{a}{b}, 
\vvote{a}{c}{p}{b} and 
\vvote{b}{p}{a}{c}. 
The total scores from non-manipulators are as follows:
$s(a)=14K + 18$,
$s(b)=s(c)=17K + 18$
and $s(p)=12K + 18$.
\myOmit
{$$
  \begin{array}{cc}
a : &12*K + 18 \\
b : &17*K + 18 \\
c : &17*K + 18 \\
d : &14*K + 18 \\
av: & 15*K + 18 \\
  \end{array}
$$
}
For each
integer $k_i$,
we have a member of the manipulating coalition
with weight $k_i$.

Now, suppose there is a solution to the {\sc partition} instance.
Let the manipulators corresponding to the integers in one half of
the partition vote
\vvote{p}{a}{b}{c}, 
and let the
others vote
\vvote{p}{a}{c}{b}.
\myOmit{
\begin{table}
\label{t:nonman}
\center{
\begin{tabular}{|c|c|c|c|c|c|c|c|}
  \hline
  weights & \multicolumn{2}{|c|}{$1^{st}$}&  \multicolumn{2}{|c|}{$2^{nd}$ }& \multicolumn{2}{|c|}{$3^{rd}$} &  {$4^{th}$}  \\
    \hline
    \multicolumn {8}{|c|}{Manipulators votes}\\
    \hline
  $K$ &$ a$ &$\succ$&$ d$ &$\succ$& $c$  &$\succ$& $b$\\
  $K$ &$ a$ &$\succ$& $d$  &$\succ$& $b$&$\succ$&$ c$ \\
  \hline
\end{tabular}
\caption{\label{t:man_2}The scores of manipulators that correspond to a valid partition}
}
\end{table}

Then the scores are as follows:
$$
  \begin{array}{cc}
a : &18*K + 18 \\
b : &18*K + 18 \\
c : &18*K + 18 \\
d : &18*K + 18 \\
av: & 18*K + 18 \\
 \end{array}
$$
} All scores are now $18K + 18$ (which is also the average). By the
tie-breaking rule, $p$ wins in the first round. Thus the
manipulators can make $p$ win if a perfect partition exists.

Conversely, suppose there is a successful manipulation. Clearly, $p$
cannot be eliminated in the first round. To ensure this, all
manipulators must put $p$ in first place. Next, we show that if $p$
is not a joint winner of the first round, $p$ cannot win overall. We
consider all possible sets of candidates that could be eliminated in
the first round. There are 6 cases. In the first case, only $a$ is
eliminated in the first round. The scores from non-manipulators in the
second round are as follows: \myOmit{
$$
  \begin{array}{cc}
a : &6*K + 10 \\
b : &12*K + 13 \\
c : &12*K + 13 \\
av: & 10*K + 12 \\\
 \end{array}
$$}
$s(b)=s(c)=12K + 13$,
and $s(p)=6K + 10$.
The average score is $10K + 12$.
Even with the maximum $4K$ possible
score from the manipulators, $p$ is eliminated.
This contradicts the assumption that $p$ wins.
In the second case, only $b$ is eliminated in the first
round. As $p$ and $a$ are not eliminated in the first round,
the manipulators have to cast votes that
put $p$ in first place and $b$ in second place.
\myOmit{
 Now the scores of non-manipulators in the second round are
as follows:
$$
  \begin{array}{cc}
a : &9*K + 13 \\
c : &12*K + 12 \\
d : &9*K + 11 \\
av: & 10*K + 12 \\
  \end{array}
$$
Then the total scores  in the second round  are
$$
  \begin{array}{cc}
a : &13*K + 13 \\
c : &12*K + 13 \\
d : &11*K + 10 \\
av: & 12*K + 12 \\
  \end{array}
$$
}
With such manipulating votes, the scores in
the second round are:
$s(a)=11K+11$, $s(c)=12K+12$ and
$s(p)=13K+13$. The average score is $12K+12$.
Hence, $a$ is eliminated. In the next round,
$p$ is eliminated as $s(p)=5K+5$,
$s(c)=7K+7$ and the average score is
$6K+6$.
This contradicts the assumption that $p$ wins.
\myOmit{
$$
  \begin{array}{cc}
a : &5*K + 5 \\
c : &7*K + 7 \\
av: & 6*K + 6 \\
  \end{array}
$$}
In the third case, only $c$ is eliminated in the first
round. This case is symmetric to the second case.
In the fourth case, $a$  and $b$ are eliminated in the first
round. The case when $a$  and $c$ are eliminated is symmetric.
 In the second round, the scores from non-manipulators are
$s(c)=7K+7$ and $s(p)=3K+5$.
\myOmit{
$$
  \begin{array}{cc}
a : &3*K + 5 \\
c : &7*K + 7 \\
av: & 5*K + 6 \\
  \end{array}
$$
}
The $2K$ score from the manipulators
cannot prevent $p$ being eliminated.
This contradicts the assumption that $p$ wins.
In the fifth case, $b$ and $c$ are eliminated
in the first round. However, in the first round,
the score $b$ and $c$ receive from the non-manipulators
is $17K + 18$. One of them will get at least $K$ points from manipulators.
This will give them greater than the average score of $18K+8$.
Hence, at least one of them is not eliminated.
In the sixth and final case, $a$, $b$ and $c$ are all eliminated in the first
round.  This case is again impossible by the same argument as the last case.

The only way for $p$ to win is to have a tie with all candidates in
the first round. As we observed above, the manipulators have to put
$p$ in first place, and $a$ in second place. In turn, both $b$ and
$c$ have to get exactly $K$ points from the manipulators. Hence,
there exists a solution to the {\sc partition} instance.
\hfill\myqed

Clearly, it is polynomial
to compute a manipulation of Baldwin's rule
with 2 candidates (since
this case degenerates to majority voting).
With Nanson's rule, on the other hand, it is
polynomial with up to 3 candidates.

\begin{mytheorem}
With Nanson's rule and weighted votes,
the coalition manipulation problem is polynomial for up to
3 candidates.
\end{mytheorem}
{\myproof Consider an election with 3 candidates ($a,b$ and
$p$) in which the manipulators want $p$ to win. We prove that the
optimal strategy is for the
manipulators either all to vote  $p\succ a\succ b$ 
or all to vote  $p\succ b\succ a$. 
If $p$ does not win using one of these two votes, then $p$ cannot win.
Therefore we simply try out the two votes and compute if $p$ wins in
either case.

Suppose the manipulators can make $p$ win. We first note that there
is no loss for them to raise $p$ to the first position, while
keeping the other parts of their preferences the same.
By doing so, the score of $p$ goes up and the scores of $a$
and $b$ go down. The only possible change in the elimination process
is that now both $a$ and $b$ drop out in the first round, so that
$p$ still wins.

Now, suppose that all manipulators rank $p$ in their top positions.
Let $P^M$ denote the manipulators' profile that makes $p$ win.
Because Nanson's rule never selects the Condorcet loser,
$p$ cannot be beaten by both $a$ and $b$ in pairwise elections.
Without loss
of generality, suppose $p$ beats $a$. We
argue that if all manipulators vote $p\succ a\succ b$, then $p$
still wins. For the sake of contradiction, suppose all manipulators
vote $p\succ a\succ b$ but $p$ does not win. As the
manipulators still rank $p$ in their top positions,
the score of $p$ in the first round is the same as in
$P^M$. Therefore, $p$ must enter (and lose) the
second round. Hence, only
$a$ is eliminated in the first round,
and in the second round $b$ beats $p$. However,
having the manipulators vote $p\succ a\succ b$ only lowers $b$'s score in
the first round, compared to the case where they vote $P^M$.
Hence, when the manipulators vote $P^M$, $b$ also enters the
second round and then beats $p$, which is a contradiction.

Therefore, if the manipulators can make $p$ win, then they can make
$p$ win by all voting $p\succ a\succ b$, or all voting $p\succ
b\succ a$. \hfill\myqed}
\section{Approximation Methods}

\newcommand{\reverse}{\mbox{\sc Reverse}\xspace}
\newcommand{\largestfit}{\mbox{\sc LargestFit}\xspace}
\newcommand{\averagefit}{\mbox{\sc AverageFit}\xspace}
\newcommand{\eliminate}{\mbox{\sc Eliminate}\xspace}
\newcommand{\reveliminate}{\mbox{\sc RevEliminate}\xspace}

One way to deal with 
computational intractability is to treat
computing a manipulation as an approximation problem where we try to
minimize the number of manipulators. 
We therefore considered five
approximation methods. These are either derived from 
methods 
used with 
Borda 
or are specifically designed
for the elimination style of Nanson's and Baldwin's rules.

\begin{description}
\item[\reverse:]
The desired candidate is put first, and the other
candidates are reverse ordered by their current Borda score. We
repeat this construction until the desired candidate wins. \reverse
was used to manipulate the Borda rule in
\cite{Zuckerman09:Algorithms}.

\item[\largestfit:]
This method was proposed for the Borda rule
\cite{dknwcomsoc10}.
Unlike \reverse which constructs votes one by
one, we 
construct votes in any order using a bin packing heuristic which
puts the next largest Borda score into the ``best'' available vote.
We start with a target number of manipulators. Simple counting
arguments will lower bound this
number, 
and we can 
increae it until we
have a successful manipulation.
We construct votes for the
manipulators in which the desired candidate
is 
in first place. We take the other Borda scores
of the manipulators in decreasing order, and
assign them to the candidate with
the lowest current Borda score who has been assigned less than the
required number of scores. A perfect matching
algorithm then converts the sets of
Borda scores for the candidates into a set of manipulating
votes.

\item[\averagefit:]
This method was also proposed for the Borda rule \cite{dknwcomsoc10}.
We again have a target number of manipulators,
and construct votes for the
manipulators in which the desired candidate
is 
in first place. We take the other Borda scores
of the manipulators in decreasing order, and
assign them to the candidate with
the current lowest {\em average} Borda score
who has less than the required number of scores.
The intuition is that if every score was of
average size, we would have a perfect fit.
If more than one candidate has the same
lowest average Borda score and can accommodate
the next score, we tie-break on the candidate with the fewest scores.
Examples of \largestfit and \averagefit can be found in \cite{dknwcomsoc10}.
\item[\eliminate:]
We repeatedly construct votes in which the desired candidate
is put in first place, and the other candidates
in the reverse of the current elimination
order. For instance, the first candidate eliminated
is put in last place. For Nanson's rule,
we order candidates eliminated
in the same round by their
Borda score in that round.

\item[\reveliminate:]
We repeatedly construct votes in which the desired candidate
is put in first place, and the other candidates
in the current elimination
order. For instance, the first candidate eliminated
is put in second place.
For Nanson's rule,
we order candidates eliminated
in the same round by the inverse of their
Borda score in that round.
\end{description}

The intuition behind \eliminate is to move the
desired candidate up the elimination order
whilst keeping the rest of the order unchanged.
With \reveliminate,
the intuition is to move the
desired candidate up the elimination order,
and to assign the largest Borda scores to the
least dangerous candidates. It is easy to show that all methods
will eventually compute a manipulation of Nanson's or Baldwin's
rule in which the desired candidate wins.

With Borda voting, good 
bounds are known on the quality
of approximation that is achievable. In particular,
\cite{Zuckerman09:Algorithms} proved that \reverse never requires
more than one extra manipulator than optimal. Baldwin's and Nanson's
rules appear more difficult to approximate within such bounds. We can
give examples where all five methods compute a manipulation that
use several more manipulators than is optimal. Indeed, even with a
fixed number of candidates, \reverse can require an unbounded number
of extra manipulators. 

\begin{mytheorem}
\label{t:path_exm_baldwin}
With Baldwin's rule,
there exists an election with
7 candidates and $42n$ votes where
\reverse computes a manipulation
with at least $n$ more votes than
is optimal.
\end{mytheorem}
\myproof (Sketch) Consider an election over $a$, $b$, $c$, $d$, $e$,
$f$ and $p$ where $p$ is the candidate that the manipulators wish to
win. We define $R(u,v)$ as the pair of votes:
$\vvote{u}{v}{\others}{p}$, $\vvote{\rev(\others)}{u}{v}{p}$ where
$\others$ is some fixed ordering of the other candidates and
$\rev(\others)$ is its reverse. The non-manipulators cast the
following votes: $3n$ copies of $R(a,b)$, $R(b,c)$, $R(c,d)$,
$R(d,e)$ and $R(e,f)$. In addition, there are $6n$ copies
of the 
votes: $\vote{p}{a}{\others}$ and $\vote{\rev(\others)}{p}{a}$. If
$18n$ manipulators vote identically $\vvote{p}{a}{\ldots}{f}$ then
$p$ wins. This provides an upper bound on
the size of the optimal manipulation.
After the non-manipulators have voted,
$s(a)=s(f)=138n$,
$s(b)=s(c)=s(d)=s(e)=141n$
and $s(p)=42n$.
\reverse will put $p$
in first place.
We suppose $n$ is a multiple of 2, but more complex arguments can be
given in other cases. After $n$ manipulating votes have been
constructed, the scores of candidates $a$ to $f$ are level at
$285n/2$ and $p$ is leveled at $48n$. From then on, the manipulators put $p$ in first place and
alternate the order of the other candidates. At least $32n$ votes are
therefore required for $p$ to move out of last place. 
\hfill\myqed

Asymptotically this result is as bad as we
could expect. Any election can be manipulated
with $O(n)$ votes by simply reversing all
previous votes, and this proof demonstrates that
\reverse may use $O(n)$ more
votes than is optimal.

\section{Experimental Results}

\newcommand{\Rev}{\mbox{\sc Rev}\xspace}
\newcommand{\LaFit}{\mbox{\sc LaFit}\xspace}
\newcommand{\AvgFit}{\mbox{\sc AvFit}\xspace}
\newcommand{\Elim}{\mbox{\sc Elim}\xspace}
\newcommand{\RevElim}{\mbox{\sc RevElim}\xspace}
\newcommand{\Opt}{\mbox{\sc nonTvl}\xspace}

\begin{table}
\begin{center}
{\scriptsize
\caption{\label{table:uni_opt} Percentage of random uniform elections
with 5 candidates
where the heuristic finds the optimal manipulation.
}
\begin{tabular}{| c||c|c|c|c|c|}
\hline
 Rules & 
\Rev & \LaFit & \AvgFit & \Elim & \RevElim  \\
\hline
\hline
Baldwin   &  74.4\% & 74.4\% & \textbf{75.8\%} & 62.2\%& 75.2\%  \\
 Nanson    &74.6\% & 76.0\% & \textbf{78.0\%} & 65.4\%& 66.9\%  \\
 Borda     & 95.7\% & 98.8\% & \textbf{99.8\%} & 95.7\%& 10.7\%  \\
     \hline
\end{tabular}}
\end{center}
\begin{center}
{\scriptsize \caption{\label{table:urn_opt} Percentage of urn
elections with 5 candidates where the heuristic finds the optimal
manipulation. }
\begin{tabular}{| c||c|c|c|c|c|}
\hline
Rules &  
\Rev & \LaFit & \AvgFit & \Elim & \RevElim  \\
\hline
\hline
Baldwin    & 75.1\% & 75.4\% & \textbf{77.3}\% & 68.9\%& 83.4\%  \\
 Nanson   & 78.1\% & 79.0\% &\textbf{ 79.8}\% & 72.2\%& 79.4\%  \\
 Borda     & 96.1\% & 92.7\% &\textbf{ 99.9}\% & 96.1\%&  4.4\%  \\
  \hline
\end{tabular}}\vspace{-4mm}
\end{center}
\end{table}

To test the difficulty of computing manipulations
in practice and the effectiveness of these approximation
methods, we ran some experiments using a similar setup
to~\cite{wecai10}.  We generated
either uniform random votes or votes drawn from a Polya
Eggenberger urn model. 
In the urn model,
votes are drawn from an urn at random, and are placed back into
the urn along with $a$ other votes of the same type.  This captures
varying degrees of social homogeneity. We set $a = m!$
so that there is a 50\% chance that the second vote is the same as the
first.

Our first set of experiments used 3000 elections with 5 candidates and
5 non-manipulating voters. This is small enough
to find the optimal number of manipulators
using brute force search, and thus to
determine how often a heuristic computes
the optimal solution. We threw out the 20\% or so of
problems generated in which the chosen candidate has already won
before the manipulators vote.
Results are given in Tables~\ref{table:uni_opt}--\ref{table:urn_opt}.
Heuristics that are very effective
at finding an optimal manipulation with the Borda rule
do not perform as well with Baldwin's and Nanson's rules.
For example,
\averagefit almost always finds an optimal manipulation
of the Borda rule but can only find an optimal solution about 3/4 of the
time with Baldwin's or Nanson's rules.

\begin{table}
\begin{center}
{\small
\caption{\label{table:uni_av_baldwin} Uniform elections using Baldwin rule.
This (and subsequent) tables give the average number of manipulators.
}
\begin{tabular}{|r|r|r|r|r|r|}
\hline
 n  & Rev     & LaFit   & AvgFit  &    Elim & RevElim \\
\hline
\hline
  4&  2.25  &  2.25  &  2.25  &  2.44  &  \textbf{2.21}  \\
  8&  \textbf{2.99}  &  3.07  &  3.01  &  3.35  &  3.06  \\
  16&  \textbf{4.31}  &  4.41  &  4.40  &  4.79  &  4.67  \\
  32& \textbf{ 5.93}  &  6.03  &  6.14  &  6.61  &  6.84  \\
  64&  \textbf{8.56}  &  8.65  &  8.84  &  9.54  &  11.02  \\
  128&  \textbf{12.13}  &  12.24  &  12.41  &  13.37  &  16.06  \\
\hline
\end{tabular}}
\end{center}
\end{table}

\begin{table}
\begin{center}
{\small
\caption{\label{table:uni_av_nanson} Uniform elections using Nanson  rule.
}
\begin{tabular}{|r|r|r|r|r|r|}
\hline
 n  & Rev     & LaFit   & AvgFit  &    Elim & RevElim \\
\hline
\hline
  4&  \textbf{2.15} &  2.17  &  2.15  &  2.25  &  2.28  \\
  8&  2.91  &  2.96  & \textbf{2.84}  &  3.05  &  3.21  \\
  16&  4.13  &  4.27  & \textbf{4.05} &  4.44  &  4.99  \\
  32&  \textbf{5.80} &  5.88  &  5.81  &  6.18  &  7.46  \\
  64&  \textbf{8.51} &  8.58  &  8.82  &  8.99  &  12.04  \\
  128&  \textbf{12.07} &  12.09  &  13.00  &  12.60  &  17.90  \\
\hline
\end{tabular}}
\end{center}
\end{table}

\begin{table}
\begin{center}
{\small
\caption{\label{table:urn_av_baldwin} Urn elections using Baldwin rule.
}
\begin{tabular}{|r|r|r|r|r|r|}
\hline
 n  & Rev     & LaFit   & AvgFit   & Elim    & RevElim \\
\hline
\hline
  4&  3.26  &  3.23  &  3.24  &  3.35  & \textbf{3.14}  \\
  8&  5.95  &  5.96  &  5.99  &  6.37  & \textbf{5.82}  \\
  16&  11.64  &  11.66  &  11.87  &  12.74  &  \textbf{11.52}  \\
  32&  \textbf{21.70}  &  21.78  &  22.35  &  24.67  &  22.41  \\
  64&  \textbf{43.09} &  43.37  &  44.24  &  49.07  &  45.70  \\
  128&  82.19  &  \textbf{81.82}  &  83.62  &  95.37  &  91.80  \\
\hline
\end{tabular}}
\end{center}
\end{table}
\begin{table}
\begin{center}
{\small
\caption{\label{table:urn_av_nanson} Urn elections using Nanson  rule.
}
\begin{tabular}{|r|r|r|r|r|r|}
\hline
 n  & Rev     & LaFit   & AvgFit   & Elim    & RevElim \\
\hline
\hline
  4&  3.20  &  \textbf{3.19} &  3.20  &  3.28  &  3.22  \\
  8& \textbf{5.93} &  5.98  &  5.95  &  6.13  &  6.09  \\
  16&  \textbf{11.62} &  11.93  &  11.64  &  12.16  &  12.37  \\
  32& \textbf{22.36}  &  22.78  &  22.53  &  24.00  &  24.39  \\
  64&  \textbf{44.56}  &  45.50  &  44.77  &  48.81  &  49.69  \\
  128&  87.18  &  87.55  & \textbf{86.76}  &  97.02  &  99.43  \\
\hline
\end{tabular}}\vspace{-5mm}
\end{center}
\end{table}

Our second set of experiments used larger problems. This amplifies
the differences between the different approximation methods (but
means we are unable to compute the optimal manipulation using brute
force search). Problems have between $2^2$ and $2^7$ candidates, and
the same number of votes as candidates. We tested 6000 instances,
1000 at each problem size.
Tables~\ref{table:uni_av_baldwin}--\ref{table:urn_av_nanson}
show the 
results for the average number of manipulators. 
The results show that overall \reverse works slightly better
than \largestfit and \averagefit , which themselves
outperform the other two methods
especially for problems with large number of candidates.
We observe a similar picture with Nanson's rule.
This contrasts with the Borda
rule where \largestfit and  \averagefit do much better
than \reverse \cite{dknwcomsoc10}.
In most cases \averagefit is  less effective
than \largestfit except urn elections with Nanson's rule.

These experimental results suggest that Baldwin's and Nanson's rules  are
harder to manipulate in practice than Borda.
Approximation methods that work well on the Borda rule are significantly less
effective on these rules. Overall, \reverse,  \largestfit and
\averagefit appear
to offer
the best performance,
though no heuristic dominates.
\vspace{-1mm}
\section{Other Related Work}\vspace{-1mm}

\citeauthor{weakestlink}
\shortcite{weakestlink}
prove that a class of
voting rules which use repeated ballots and eliminate
one candidate in each round are Condorcet consistent.
They illustrate this class with
the {\em weakest link} rule in which the candidate
with the fewest ballots in each round is eliminated.
\citeauthor{geller}
\shortcite{geller}
has proposed a variant of single transferable vote
where 
first place votes,
candidates are successively eliminated based on their
{\em original} Borda score. 
Unlike Nanson's
and Baldwin's rules, this method does not recalculate the Borda
score based on the new reduced set of candidates.
For any Condorcet consistent rule (and thus for
Nanson's and Baldwin's rule),
\citeauthor{bbhhaaai10}
\shortcite{bbhhaaai10}
showed that many types of
control and manipulation are polynomial
to compute when votes are single peaked.

\section{Conclusions}

With unweighted votes, we have proven that Nanson's and Baldwin's
rules are NP-hard to manipulate with one manipulator. This increases
by two thirds the number of rules known to be NP-hard to manipulate
with just a single manipulator. With weighted votes, on the other
hand, we have proven that Nanson's rule is NP-hard to manipulate
with just a small number of candidates and a coalition of
manipulators. We have also proposed a number of approximation
methods for manipulating Nanson's and Baldwin's rules. Our
experiments suggest that both rules are difficult to manipulate in
practice. There are many other interesting open questions coming
from these results. For example, are there other elimination style
voting rules which are computationally difficult to manipulate? As a
second example, with Nanson's and Baldwin's rule what is the
computational complexity of other types of control like the
addition/deletion of candidates, and the addition/deletion of
voters? As a third example, we could add elimination rounds to other
scoring rules. Do such rules have interesting computational
properties?

\vspace{-2mm}
\section*{Acknowledgments}
Nina Narodytska is supported
by the Asian Office of Aerospace Research and Development
through grant AOARD-104123.
Toby Walsh is funded by the Australian Department of Broadband,
Communications and the Digital Economy and the ARC. Lirong Xia
acknowledges a James B.~Duke Fellowship and Vincent Conitzer's NSF
CAREER 0953756 and IIS-0812113, and an Alfred P.~Sloan fellowship
for support. We thank all AAAI-11 reviewers for their helpful
comments and suggestions. \vspace{-2mm}


%


\end{document}